\documentclass[runningheads]{llncs}

\usepackage[T1]{fontenc}

\usepackage{booktabs}
\usepackage{comment}
\usepackage{multirow}
\usepackage{graphicx}
\usepackage{amsmath}
\usepackage{bbm}
\usepackage{cite}
\usepackage{subcaption}
\usepackage[misc]{ifsym}

\begin{document}

\title{Waypoint Generation in Row-based Crops with Deep Learning and Contrastive Clustering}

\titlerunning{Waypoint Generation in Row-based Crops with Contrastive Clustering}

\author{Francesco Salvetti\inst{1,2,3} (\Letter)\and
Simone Angarano\inst{1,2}\and
Mauro Martini\inst{1,2}\and \newline
Simone Cerrato\inst{1,2}\and
Marcello Chiaberge\inst{1,2}}

\authorrunning{F. Salvetti et al.}

\institute{Politecnico di Torino, Turin, Italy \\
\email{name.surname@polito.it} \and
PIC4SeR (PoliTo Interdepartmental Centre for Service Robotics), Turin, Italy \and
SmartData@PoliTo, Turin, Italy}

\toctitle{Waypoint Generation in Row-based Crops with Deep Learning and Contrastive Clustering}
\tocauthor{Francesco~Salvetti}

\maketitle 

\begin{abstract}
The development of precision agriculture has gradually introduced automation in the agricultural process to support and rationalize all the activities related to field management. In particular, service robotics plays a predominant role in this evolution by deploying autonomous agents able to navigate in fields while executing different tasks without the need for human intervention, such as monitoring, spraying and harvesting. In this context, global path planning is the first necessary step for every robotic mission and ensures that the navigation is performed efficiently and with complete field coverage. In this paper, we propose a learning-based approach to tackle waypoint generation for planning a navigation path for row-based crops, starting from a top-view map of the region-of-interest. We present a novel methodology for waypoint clustering based on a contrastive loss, able to project the points to a separable latent space. The proposed deep neural network can simultaneously predict the waypoint position and cluster assignment with two specialized heads in a single forward pass. The extensive experimentation on simulated and real-world images demonstrates that the proposed approach effectively solves the waypoint generation problem for both straight and curved row-based crops, overcoming the limitations of previous state-of-the-art methodologies.
\end{abstract}

\section{Introduction}
Agriculture 4.0 is introducing digital tools and technologies in precision farming. According to this innovative paradigm, Big Data, Artificial Intelligence and robotics play a key role in increasing the economic, environmental and social sustainability of agricultural processes, thanks to the efficient and automatic data collection and the processing tools they provide. In the last years, Deep Learning (DL) research has been substantially contributing to the development of new technologies for precision agriculture applications \cite{zhai2020decision,unal2020smart,kamilaris2017review,kamilaris2018deep}. In particular, several computer vision techniques have proven effective in supporting visual agricultural tasks such as fruit detection and counting \cite{mazzia2020real}, plant disease identification \cite{mohanty2016using,ferentinos2018deep}, land coverage and vegetation index classification from satellite and Unmanned Aerial Vehicles (UAVs) imagery \cite{khaliq2019refining,martini2021domain,mazzia2020uav}.

Besides image processing systems, self-driving machines represent a crucial component to reduce the costs of agricultural processes by providing autonomous, full-time and weather-independent operators. In this regard, the confidence in solutions based on autonomous Unmanned Ground Vehicles (UGVs) and UAVs is drastically increasing \cite{sparrow2021robots,tripicchio2015towards}, considering the competitive advantage they can provide supporting tasks such as crop monitoring \cite{monitoring}, spraying \cite{deshmukh2020design}, grasping \cite{kang2020real}, and harvesting \cite{luo2018vision}. 
In this context, designing a reliable autonomous navigation system in constrained row-based crops such as vineyards and orchards is fundamental. So far, many works proposed local planners combining deep learning with computer vision \cite{aghi2020autonomous,aghi2020local} or other sensor processing methods \cite{riggio2018low,astolfi2018vineyard,barawid2007development}. However, local planners provide a solution for intra-row navigation only, and therefore a global path generator is always needed. In a complex scenario such as a row-based environment, where traversing each row is the practical navigation goal, the problem of developing an efficient global path planner has been quite neglected by the research community. Existing solutions usually tackle the problem by clustering visual data obtained from satellites or UAVs. For example, in \cite{zoto2019automatic} authors use a classical clustering method to identify vineyard rows from a 3D model of the terrain reconstructed from UAV data and then compute the path accordingly. However, as pointed out in \cite{vidovic2014center}, the extraction of relevant information about rows geometry from images can be a complex task, in addition to being extremely computationally expensive. This limitation also holds considering other approaches besides clustering. For instance, in \cite{comba2018unsupervised} authors adopted 3D point cloud aerial photogrammetry to detect the structure of vineyards.

Recently, the DeepWay method \cite{mazzia2021deepway} has been proposed to efficiently combine deep learning and clustering for the generation of start and end row waypoints given an occupancy grid of the vineyard. Moreover, novel contributions adopted the same paradigm and training procedure to extend the coverage to arbitrary unstructured environments \cite{lei2022deep}. Despite being an important baseline for row-based path generation, DeepWay leaves substantial space for improvement. In particular, it applies DBSCAN clustering \cite{ester1996density}, followed by a complex heuristic geometrical post-processing heavily based on angle estimation, to discriminate start waypoints from end waypoints. However, this method performs poorly in a wide range of real-world situations, including curved crops and rows of different lengths.

In this work, we propose a novel solution for waypoint generation in row-based crops, combining deep learning with a contrastive clustering approach. To this end, we conceive a new DNN architecture to simultaneously predict the position of the navigation waypoints for each row and cluster them in a single forward step. Hence, we train our model with an additional contrastive loss on a synthetic dataset of top-view vineyard maps and test it on manually-labeled real satellite images. Our extensive experimentation demonstrates that the proposed solution successfully predicts precise waypoints also in real-world crop maps. We also consider complex conditions such as curved rows, differently from previous solutions based on classical clustering algorithms. 

The contributions of this work can be therefore summarized as follows:

\begin{itemize}
    \item we propose a novel deep learning model to simultaneously predict the position of navigation waypoints and cluster them in a unique forward step;
    \item we solve the limitations of classical clustering methodologies by adopting a contrastive loss function;
    \item we present a method for synthetic generation of realistic curved occupancy grids for row-based crops;
    \item we demonstrate that our model trained on synthetic data successfully generalizes to challenging real-world satellite images.
\end{itemize}

The article is organized in the following sections. In Sec. \ref{sec:methodology}, we introduce our methodology, describing the backbone design, the waypoint estimation and the contrastive clustering separately. In Sec. \ref{sec:experimental}, we propose a thorough description of our experimentation, defining the settings and procedures used to generate the synthetic dataset and train the DNN. In Sec. \ref{sec:results}, we report and discuss the obtained results, comparing our solution with classical clustering baselines, and, in Sec. \ref{sec:conclusions}, we draw some conclusions.

\section{Methodology}
\label{sec:methodology}
 Due to its intrinsic nature, every row-based crop is characterized by a set of lines or curves that identify two regions comprising the starting and ending points of each row, respectively. In this scenario, a robotic path should cover the whole field, and it can be divided into intra-row segments, that connect the starting region to the ending region, and inter-row segments, that connect two starting or two ending points. Given an optimal estimation of these starting and ending waypoints, it is possible to plan a full-coverage path in the row-based environment simply by alternating intra-row and inter-row segments. Therefore, the planning process heavily relies on two main steps: waypoint estimation, which identifies candidates for the points of interest, and waypoint clustering, which assigns each estimated point to one of the two regions.
 
 Following the same approach presented in \cite{mazzia2021deepway}, we frame the waypoint generation process as a regression problem, in which we estimate the coordinates of the points with a deep neural network, starting from a top-view map of the environment. The map consists of a 1-bit single-channel occupancy grid that identifies with 1 the plant rows and with 0 the free terrain. Therefore, this kind of estimation process can be easily applied to geo-referenced segmented masks of the target fields obtained from satellites or UAV imagery. The waypoints and the planned path can then be converted from the image reference system to a Global Navigation Satellite System (GNSS) reference frame to be used in real-world navigation. In addition to waypoint detection, differently from classical unsupervised methodologies for point clustering, we propose a supervised approach based on a contrastive loss to perform point assignment. Therefore, the proposed model simultaneously performs both estimation and clustering with a single forward pass, without the need for complex post-processing operations based on heuristic geometrical-based rules.
 
\begin{figure}[t]
\centering
\includegraphics[trim={0 0 0 0}, clip, width=\textwidth]{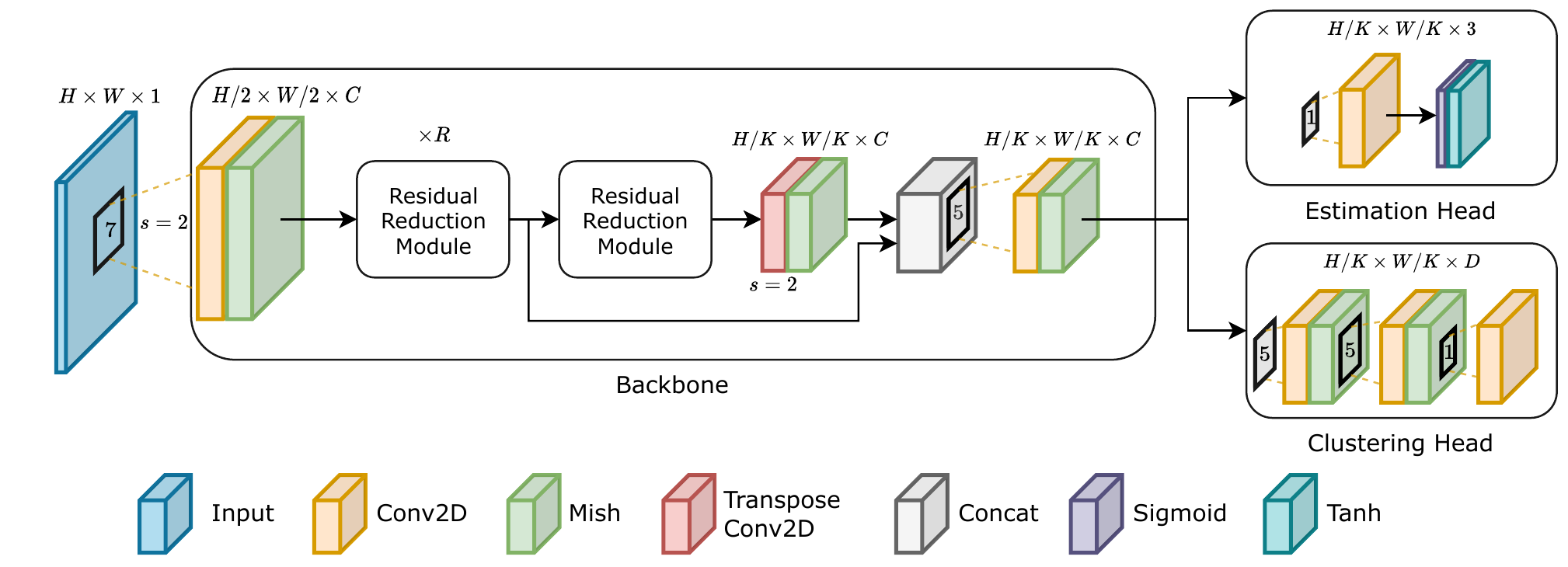}
\caption{Architecture of the backbone and the two regression heads. The number of residual reduction modules in the main block $R$ determines the backbone compression factor $K = 2^{R+1}$.}
\label{fig:architecture}
\end{figure}

\begin{figure}[t]
\centering
\includegraphics[trim={0 0 0 0}, clip, width=\textwidth]{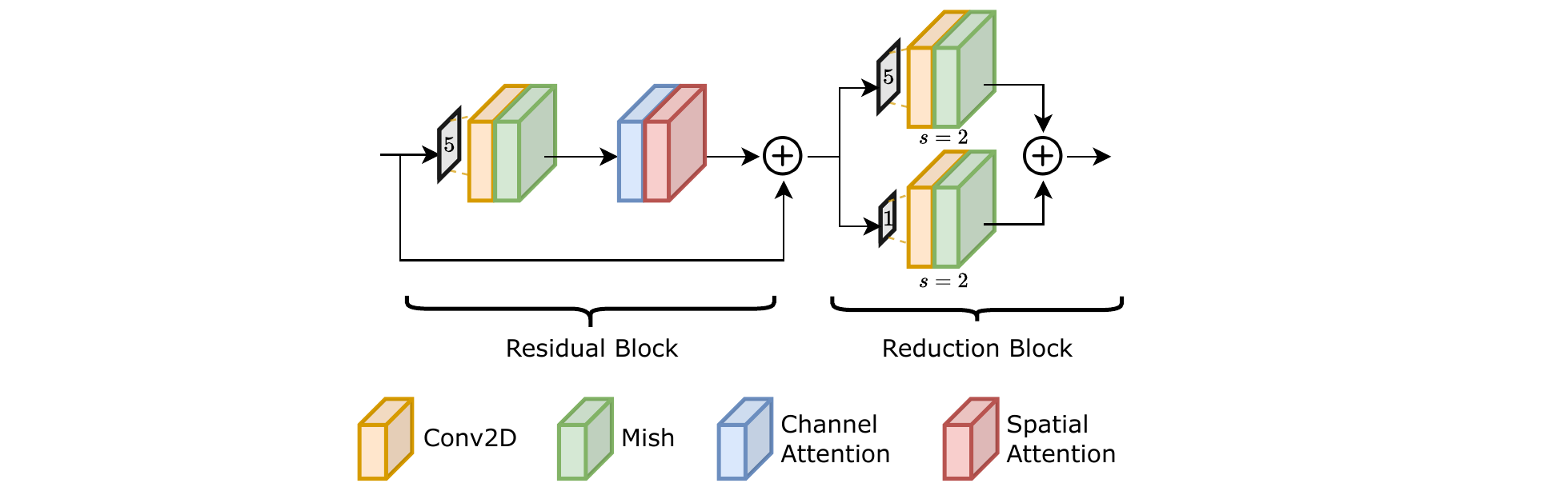}
\caption{Residual reduction module architecture. The channel and spatial attentions are implemented as in \cite{woo2018cbam}.}
\label{fig:resared}
\end{figure}

\subsection{Backbone Design}
\label{subsec:backbone}
We implement the model as a convolutional neural network characterized by a feature extraction backbone, followed by two specialized heads. A head is responsible for the estimation task, while the other deals with clustering.

The backbone is designed following the same architecture used in \cite{mazzia2021deepway}. The basic block of the network is the residual module, characterized by a stack of a 2D convolution and spatial and channel attention \cite{woo2018cbam}. Each residual block is followed by a reduction module characterized by convolutions with stride 2 that progressively halve the spatial dimensions. The backbone is a stack of $R$ residual reduction modules, made by combining a residual module and a reduction module. The final part of the network is made by an additional downsampling block, followed by a transposed convolution upsampling stage, all arranged in a residual fashion. This combination of compression and expansion has been proved very effective for different computer vision tasks such as segmentation \cite{ronneberger2015u} and representation learning \cite{tschannen2018recent}. Overall, the model performs a dimensionality compression of a factor of $2^{R+1}$, where $R$ is the number of residual reduction modules in the main block. The complete backbone structure is detailed in Fig. \ref{fig:architecture} and Fig. \ref{fig:resared}.

\subsection{Waypoint Estimation}
\label{subsec:waypoints}
\begin{figure}[t]
\centering
\includegraphics[trim={90pt 10pt 0 0}, clip, width=0.7\textwidth]{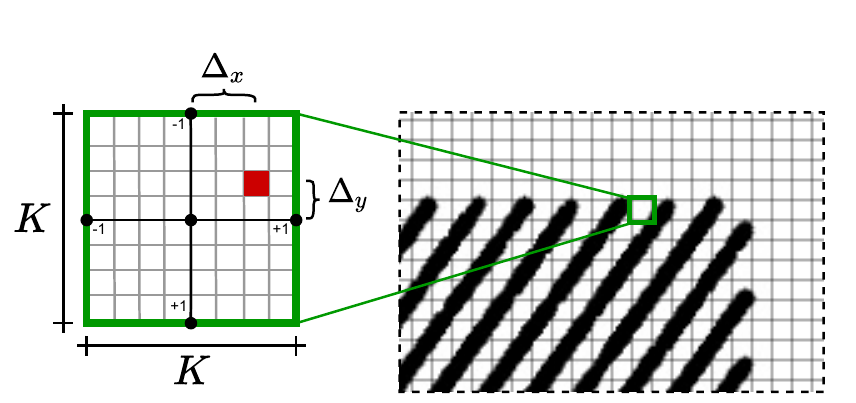}
\caption{The input occupancy grid is subdivided into a grid of $K\times K$ cells. For each cell, the waypoint estimation head outputs the probability $p$ of a waypoint presence, as well as the relative horizontal and vertical displacements with respect to the cell center $\boldsymbol{\Delta} = (\Delta x, \Delta y)$. }
\label{fig:grid}
\end{figure}

The waypoint estimation is framed as a regression problem, similarly to object detection approaches in computer vision \cite{redmon2016you}. In particular, given an input occupancy grid map $X$ with dimensions $H\times W$, we subdivide it into a grid of $K\times K$ cells. Each cell is responsible for predicting the probability $p$ that a waypoint falls in that region, as well as its relative horizontal and vertical displacements with respect to the cell center $\boldsymbol{\Delta} = (\Delta x, \Delta y)$. The displacements are defined in the range $[-1,+1]$ and represent a shift relative to half of the cell dimension, with $-1$ identifying the left/top borders and $+1$ the right/bottom ones. An example of prediction with its correspondent displacements is shown in Fig. \ref{fig:grid}. Given a prediction $\boldsymbol{\hat{p}_{out}} = (\hat{x}_{out}, \hat{y}_{out})$ in the output reference frame, the waypoint coordinates in the input reference frame $\boldsymbol{\hat{p}_{in}}$ can be reconstructed with the following equation:

\begin{equation}
\boldsymbol{\hat{p}_{in}} = \boldsymbol{\hat{p}_{out}}\;K + \frac{K}{2} + \boldsymbol{\Delta}\frac{K}{2}
\end{equation}

The waypoint estimation head maps the high-level features extracted with the backbone to the output space with a 1x1 convolution. The backbone compression factor $2^{R+1}$ corresponds to the grid dimension $K$. Therefore, the output tensor of the estimation branch has a dimension of $H/K\times W/K\times 3$. We apply a sigmoid activation to the probability output and a tanh activation to the displacement outputs. We optimize the network for the waypoint estimation task with a weighted mean squared error loss. For each output cell $\boldsymbol{u_{i,j}}$, the estimation loss is therefore computed as:

\begin{equation}
\label{eq:estimation_loss}
    l_{i,j}^{\text{\;est}}=
    \mathbbm{1}^\text{wp}_{i,j}\lambda
    \lVert\boldsymbol{u_{i,j}}-\boldsymbol{\hat{u}_{i,j}}\rVert_2 +
    (1-\mathbbm{1}^\text{wp}_{i,j})(1-\lambda)
    \lVert\boldsymbol{u_{i,j}}-\boldsymbol{\hat{u}_{i,j}}\rVert_2
\end{equation}

\noindent where $\mathbbm{1}^\text{wp}_{i,j}\in\{0,1\}$ is an indicator Boolean function evaluating 1 if a waypoint is present in that cell, and $\lambda$ is the relative constant that weights differently positive and negative cells.

At inference time, we get the list of predicted waypoints by considering all the cells with probability $p$ over a certain threshold $t_p$. As in standard object detection methodologies, we also apply a suppression algorithm to decrease the number of redundant predictions that typically occur when multiple adjacent cells detect the same waypoint. The algorithm identifies all the groups of predictions with Euclidean distance within a certain threshold $t_{sup}$ in the input reference frame. For each group, the point with highest confidence $p$ is selected, while the remaining predictions are discarded. 

\begin{figure}[t]
\centering
\includegraphics[trim={0 0 0 0}, clip, width=\textwidth]{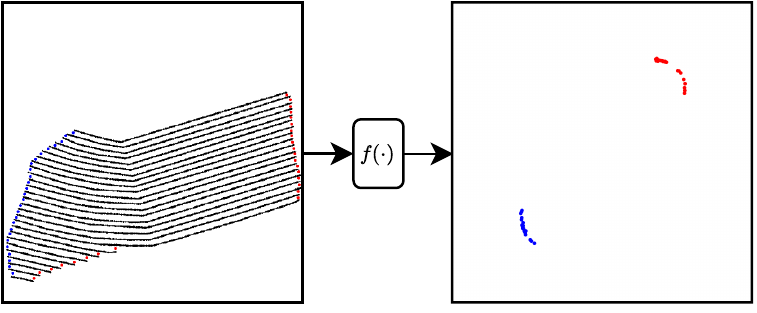}
\caption{In the latent space mapped by $f(\cdot)$, points of the same cluster appear closer together with respect to points of the other cluster. The mapping function $f(\cdot)$ is implemented with the backbone and the clustering head together. In this example, the latent space has a dimensionality $D=2$.}
\label{fig:clustering}
\end{figure}

\subsection{Contrastive Clustering}
\label{subsec:clustering}
Once the waypoints are detected, they should be assigned to starting or ending regions. This task can be seen as a simple binary classification, in which the labels represent the two clusters. However, in this scenario the actual assigned label is not relevant, as the only fundamental aspect is whether points of the same group are assigned the same label. The aim is to discriminate the points of the two regions without caring about which of them is classified as starting or ending. Indeed, an optimal path can be successfully planned regardless of the choice of the starting cluster. This invariance cannot be guaranteed by supervised classification.

For this reason, we model the clustering problem as a supervised representation learning process. Given the two sets of points $A = \{\boldsymbol{p}\mid \boldsymbol{p}\in\text{first cluster}\}$ and $B = \{\boldsymbol{p}\mid \boldsymbol{p}\in\text{second cluster}\}$, we want to find a non-linear mapping $f(\cdot)$ such that

\begin{equation}
    d\big(f(\boldsymbol{p_i}), f(\boldsymbol{p_j})\big) \ll d\big(f(\boldsymbol{p_i}), f(\boldsymbol{p_k})\big) \quad
    \text{for} \quad \boldsymbol{p_i},\boldsymbol{p_j} \in A \;,\; \boldsymbol{p_k} \in B
\end{equation}

 \noindent and vice versa, where $d$ is a distance measure. In the latent space mapped by $f(\cdot)$, points of different clusters are well-separated according to distance $d$. This means that a simple clustering method such as K-means \cite{macqueen1967some} can successfully discriminate the two groups in the latent space, as shown in Fig. \ref{fig:clustering}. Inspired by the contrastive framework used for unsupervised learning in \cite{chen2020simple}, we select as distance metric $d$ the inverse of the cosine similarity:
 
\begin{equation}
    \text{sim}(\boldsymbol{u},\boldsymbol{v}) = \frac{\boldsymbol{u}^\top\boldsymbol{v}} {\lVert\boldsymbol{u}\rVert_2\;\lVert\boldsymbol{v}\rVert_2}
\end{equation}

For each image, we consider the $N$ ground-truth waypoints as independent samples. Given a point $\boldsymbol{p_i}$, we consider as positive examples all the other $N/2 -1$ points in the same cluster, and as negative examples the $N/2$ points of the other cluster. Therefore, we define the clustering loss contribution for the sample $i$ as:
\begin{equation}
\label{eq:clustering_loss}
\begin{split}
l_i^{\text{\;clus}} = \frac{1}{N-1}\sum_{\substack{j=1 \\ j\neq i}}^N \; \bigg[\; \quad
&\text{\Large$\mathbbm{1}$}_{\substack{
                        \!\!\!\!\boldsymbol{p_i},\boldsymbol{p_j} \in A\\
                        \vee\,  \boldsymbol{p_i},\boldsymbol{p_j}\in B}}
\;\log\Big(\text{sig}\,\big(
           \text{sim}\,\big(f(\boldsymbol{p_i}),f(\boldsymbol{p_j})\big)
            \big)\Big)  \; + \\ +\;
\Big( 1 - &\text{\Large$\mathbbm{1}$}_{\substack{
                        \!\!\!\!\boldsymbol{p_i},\boldsymbol{p_j}\in A\\
                        \vee\,  \boldsymbol{p_i},\boldsymbol{p_j}\in B}}
\,\Big) \;\log\Big(1-\text{sig}\,\big(
              \text{sim}\,\big(f(\boldsymbol{p_i}),f(\boldsymbol{p_j})\big)
              \big)\Big) \bigg]
\end{split}
\end{equation}

\noindent where $\text{\Large\(\mathbbm{1}\)}_{\substack{
                    \!\!\!\!\boldsymbol{p_i},\boldsymbol{p_j}\in A\\
                    \vee\,  \boldsymbol{p_i},\boldsymbol{p_j}\in B\\\,}}
\in \{0,1\}$ is an indicator function evaluating 1 if $\boldsymbol{p_i}$ and $\boldsymbol{p_j}$ are in the same cluster and 0 otherwise, while `sig' represents the sigmoid function. Basically, this loss computes the binary cross-entropy of the cosine similarity in the latent space mapped by $f(\cdot)$ for the pair $(\boldsymbol{p_i},\boldsymbol{p_j})$. $f(\cdot)$ is optimized to push the cosine similarity towards the maximum +1 if the points are in the same cluster and towards the minimum -1 otherwise. The final loss is computed over all the pairs $(i,j)$ as well as $(j,i)$ for each input image. This loss can be seen as a variation of the one used in \cite{wu2018unsupervised,van2018representation,chen2020simple}, but instead of $N$ groups with 2 elements each, optimized with categorical cross-entropy and softmax, we have 2 groups with $N/2$ elements each, optimized with binary cross-entropy and sigmoid.

The mapping $f(\cdot)$ is modeled by the clustering head in the output space reference system. The head is composed of two convolutional layers with Mish activation and one final 1x1 convolution with linear activation. The output tensor of the clustering branch has a dimension of $H/K\times W/K\times D$, where $D$ is the latent space dimensionality.

At inference time, for each waypoint detected in the estimation phase, we select the correspondent feature from the clustering head output. We can predict the clustering assignment by fitting a K-means predictor with two centroids on the selected features. Since we use the cosine similarity in the loss computation, we are optimizing the clustering in the normalized latent space. For this reason, the features should be divided by their Euclidean norm before clustering. This normalization decreases by one the latent space dimensionality, and therefore the minimum number of dimensions $D$ for the clustering head is 2.  

\section{Experimental Setting}
\label{sec:experimental}
In this section, we present all the details of our experimentation. We describe the datasets used for network training and testing as well as the main hyperparameters adopted during the training phase.

\begin{figure}[t]
\centering
\begin{subfigure}{0.3\textwidth}
  \centering
  \frame{\includegraphics[trim={50pt 50pt 20pt 20pt},clip,width=\linewidth]{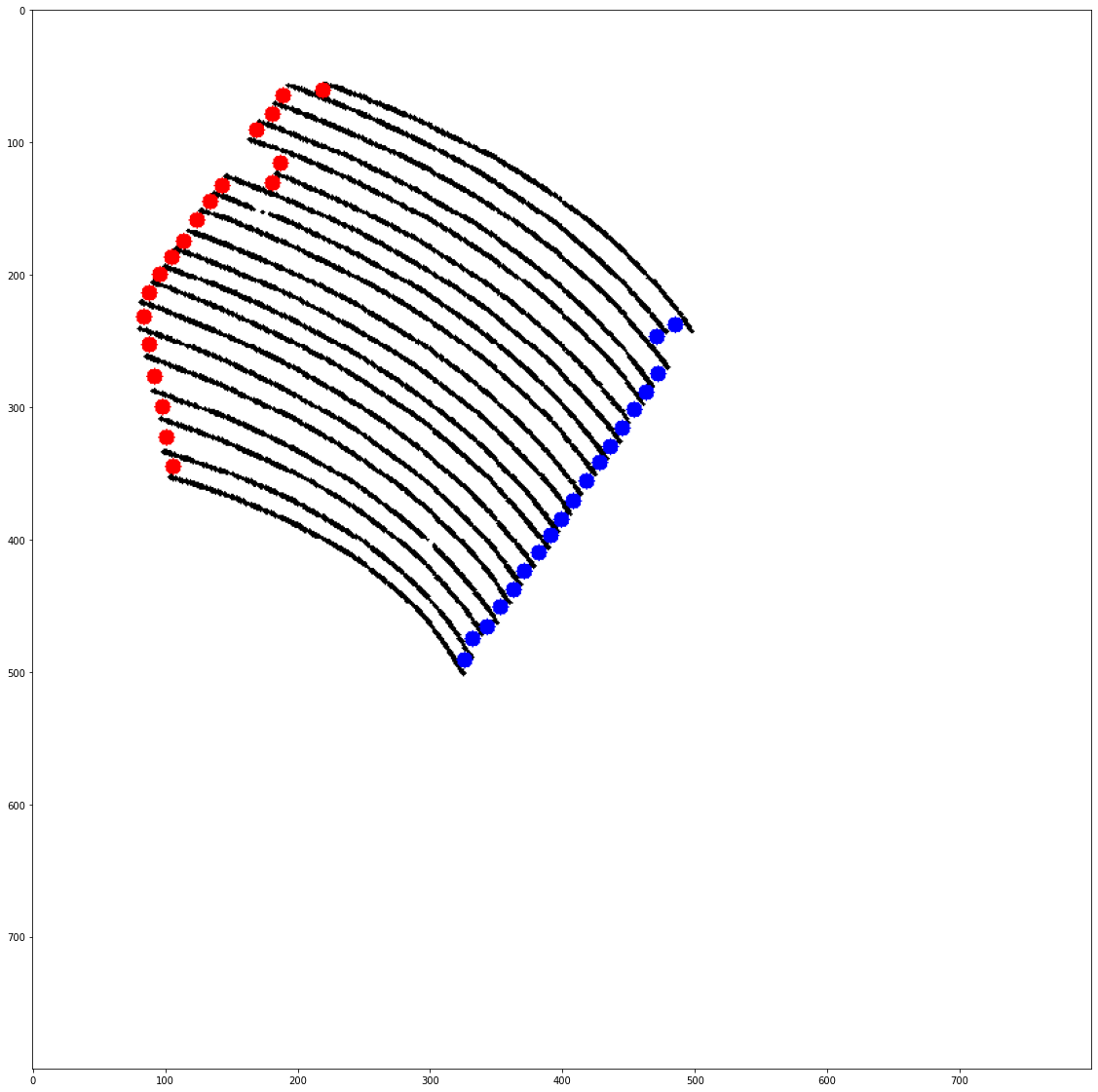}}
  \caption{}
\end{subfigure}
\begin{subfigure}{0.3\textwidth}
  \centering
  \frame{\includegraphics[trim={50pt 50pt 20pt 20pt},clip,width=\linewidth]{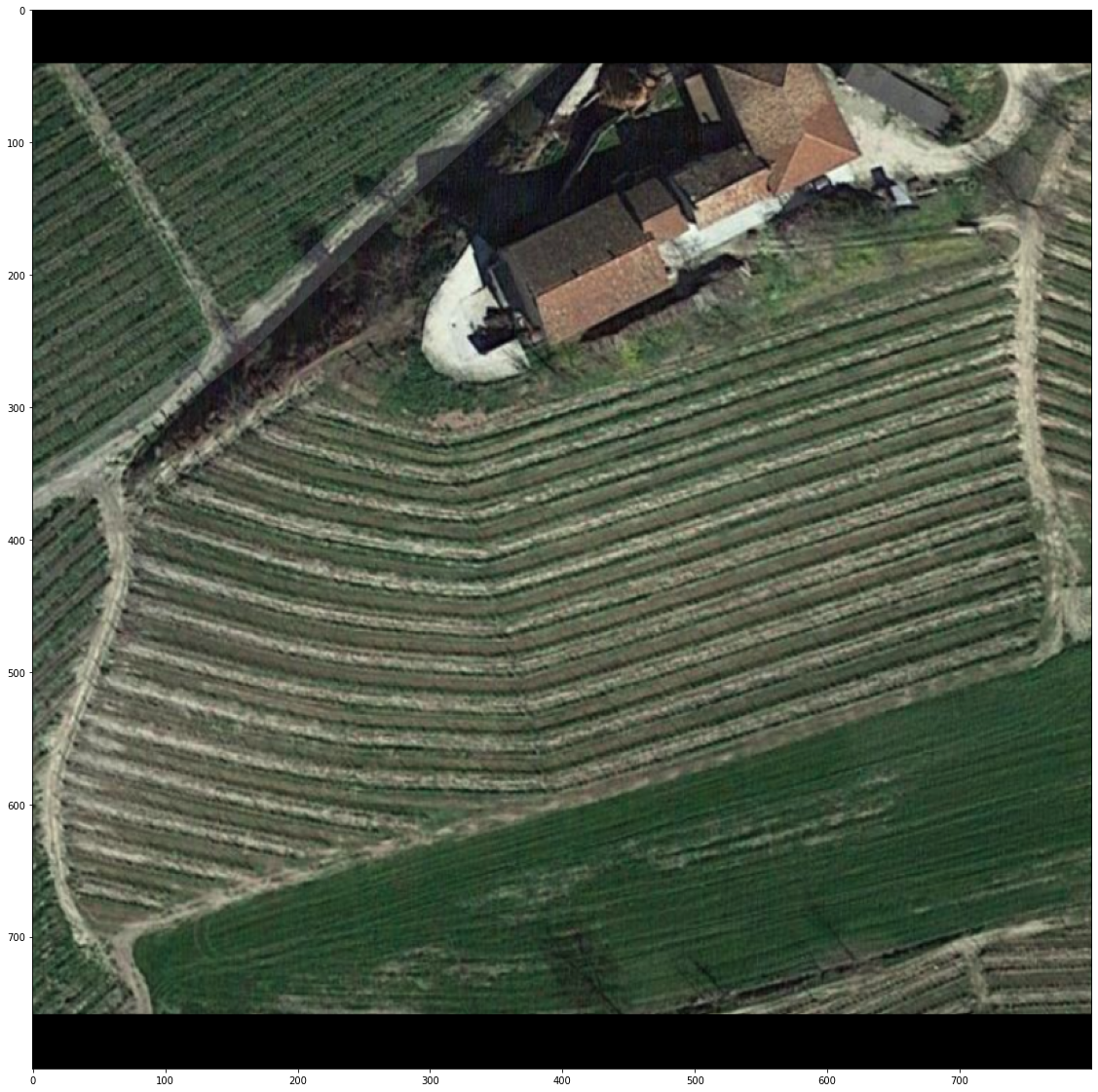}}
  \caption{}
 \end{subfigure}
\begin{subfigure}{0.3\textwidth}
  \centering
  \frame{\includegraphics[trim={50pt 50pt 20pt 20pt},clip,width=\linewidth]{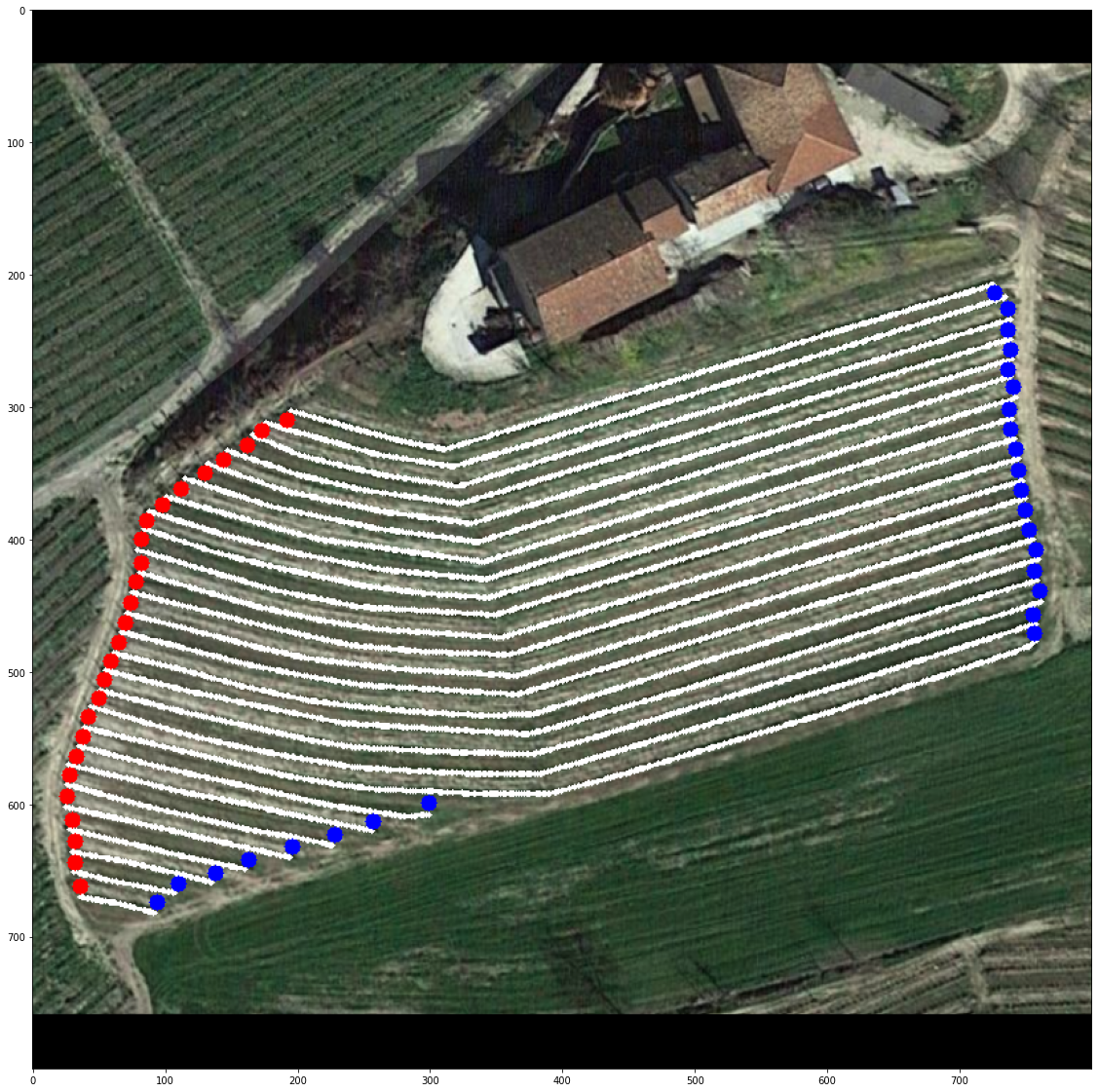}}
  \caption{}
 \end{subfigure}
\caption{Examples of curved occupancy grids: synthetic (a) and real-world from Google Maps satellite database without (b) and with (c) manual annotation. Red and blue points are the ground-truth waypoints divided in the two clusters.}
\label{fig:dataset}
\end{figure}

\subsection{Dataset Description}
\label{subsec:dataset}
Considering the lack of open datasets of row crops bird-eye maps and the time required to manually annotate a large set of real images, we define a method to build realistic synthetic occupancy grids to train the model. We modify the method presented in \cite{mazzia2021deepway} to extend it to both straight and curved occupancy grids. The generation process can be summarized as follows:

\begin{enumerate}
    \item sample a uniformly random number of rows $n\in[10,50]$ and angle $\alpha\in[-\pi/2,\pi/2]$;
    \item generate row centers with a random inter-row distance, along the line perpendicular to $\alpha$ and passing through the image center;
    \item generate random field borders and find starting and ending points for each row with orientation $\alpha$;
    \item to create curved maps, add a random displacement to the row centers and compute a quadratic Bézier curve with the starting, ending and center points as control points; this ensures that the curves are continuous and smooth;
    \item generate the occupancy grid by drawing circles with random radius $r\in[1,2]$ pixels to model irregularities in the row width
    \item create random holes in the rows to emulate segmentation errors or missing plants;
    \item compute the $N=2n$ ground-truth waypoints as the mean points of the lines connecting the ending points of the rows with the adjacent ones.
\end{enumerate}

To further increase variability, we randomly add displacement noise every time we sample a point coordinate during the generation process. We select $H=W=800$ pixels as input dimension for all the generated images. To investigate the effect of including synthetic curved images in the training set, we randomly generate two independent datasets, one with straight rows only, the other with both straight and curved rows. Overall, each dataset contains 3000 images for training, 300 for validation, and 1000 for testing. In addition to the synthetic data, we manually annotate real row-based images of vineyards and orchards from Google Maps (100 straight and 50 curved). These satellite images are fundamental to test the ability of the network to generalize to real-world scenarios and to prove the effectiveness of the synthetic generation process. Figure \ref{fig:dataset} shows examples of both synthetic and manually-annotated images.

\subsection{Network Training}
\label{subsec:training}

To select the best hyperparameters, we perform a random search over a set of reasonable values. For all the convolutional layers, we set a kernel size of 5 and channel dimension $C=16$. For the main block of the backbone, we set the number of residual reduction modules $R=2$. Therefore, the backbone compression factor and output cell dimension is  $K = 2^{(R+1)} = 8$. We set the clustering space dimensionality to $D=3$. Thus, the output tensors have both a dimension of $100\times100\times3$. The resulting network is a lightweight model with less than 73,000 parameters. We select Adam \cite{kingma2014adam} as optimizer with a constant learning rate of $\eta=3\mathrm{e}{-4}$ and batch size of 16. Experimentally, we find more effective to first train the estimation head and the backbone together with the loss of Eq. \ref{eq:estimation_loss}. We set the loss weight to $\lambda = 0.7$ to compensate for the high imbalance in the number of positive and negative cells and stabilize the training. We then freeze the backbone weights and train the clustering head only with the loss of Eq. \ref{eq:clustering_loss}. To highlight the challenge posed by curved scenarios, we independently train the model on both the straight and curved training sets. We train each model for a total of 200 epochs on an Nvidia 2080 Ti GPU using the TensorFlow 2 framework. To obtain significant statistics, we run each training session three times, so that the results can be described in terms of mean and standard deviation.

\begin{table}[t]
\centering
\caption{Performance of waypoint estimation on both straight and curved test datasets. We first test the model on our synthetic datasets (Straight Synth, Curved Synth) and then validate the results on manually annotated occupancy grids obtained from real satellite images (Straight Real, Curved Real). For each test set, we compare the results of the model trained on straight rows with those obtained training on curved rows. We report the mean and standard deviation for the Average Precision $AP_r$, where $r$ is the maximum accepted distance in pixels between predicted and ground-truth waypoints.}\label{tab:mask}
\resizebox{\textwidth}{!}{%
\begin{tabular}{@{}clclclclclclc@{}}
\toprule
\textbf{Test}                  &  & \textbf{Train} &  & \boldmath$AP_2$   &  & \boldmath$AP_3$   &  & \boldmath$AP_4$   &  & \boldmath$AP_6$   &  & \boldmath$AP_8$    \\ \midrule \midrule
\multirow{2}{*}{Straight Synth}      &  & Straight       &  & 0.6404 ± 0.0171 &  & 0.9284 ± 0.0088 &  & 0.9856 ± 0.0021 &  & 0.9991 ± 0.0001 &  & 0.9993 ± 0.0001 \\
 &  & Curved&  & 0.5751 ± 0.0241 &  & 0.8921 ± 0.0107 &  & 0.9743 ± 0.0022 &  & 0.9979 ± 0.0001 &  & 0.9984 ± 0.0001 \\ \midrule
\multirow{2}{*}{Straight Real} &  & Straight       &  & 0.5191 ± 0.0288 &  & 0.8155 ± 0.0109 &  & 0.9116 ± 0.0032 &  & 0.9482 ± 0.0017 &  & 0.9507 ± 0.0024  \\
 &  & Curved&  & 0.4597 ± 0.0166 &  & 0.7634 ± 0.0076 &  & 0.8788 ± 0.0089 &  & 0.9391 ± 0.0052 &  & 0.9433 ± 0.0049 \\ \midrule \midrule
\multirow{2}{*}{Curved Synth}        &  & Straight       &  & 0.5143 ± 0.0193 &  & 0.8224 ± 0.0236 &  & 0.9232 ± 0.0166 &  & 0.9726 ± 0.0078 &  & 0.9768 ± 0.0065  \\
 &  & Curved&  & 0.5664 ± 0.0226 &  & 0.876 ± 0.0066  &  & 0.9632 ± 0.0009 &  & 0.9937 ± 0.0006 &  & 0.9949 ± 0.0006 \\ \midrule
\multirow{2}{*}{Curved Real}   &  & Straight       &  & 0.4685 ± 0.0906 &  & 0.7110 ± 0.0625 &  & 0.8125 ± 0.0625 &  & 0.8802 ± 0.0374 &  & 0.8891 ± 0.0355  \\
 &  & Curved&  & 0.5327 ± 0.0269 &  & 0.8010 ± 0.0095 &  & 0.8881 ± 0.0094 &  & 0.9333 ± 0.0026 &  & 0.9374 ± 0.0033 \\ \bottomrule
\end{tabular}%
}
\end{table}

\section{Results}
\label{sec:results}
In this section, we report and comment the main results regarding both waypoint detection and clustering. Visual examples are included as well, to give a qualitative idea of the performance of our model. We extensively test our approach on both straight and curved rows, including a final evaluation on real satellite data. All the related code is open source and available online\footnote{\url{www.github.com/fsalv/ClusterWay}}.

\subsection{Waypoint Estimation}

As regards waypoint estimation, we use Average Precision ($AP_r$) as principal metric, considering different values of the range threshold $r$, such that a waypoint is considered correctly detected if its Euclidean position error in pixels is smaller than $r$. In this way, we can highlight the precision of the model at different levels of proximity. The AP is commonly used for evaluating object detection tasks \cite{everingham2010pascal,lin2014microsoft} and is computed as the area-under-the-curve of the precision-recall plot obtained varying the confidence threshold $t_p$. The waypoint estimation results are reported in Table \ref{tab:mask}, where each value is detailed with its mean and standard deviation. All the tests are performed setting a waypoint suppression threshold equal to the minimum inter-row distance of the synthetic datasets, $t_{sup} = 8$ pixels.

\begin{table}[!ht]
\centering
\caption{Performance of waypoint clustering on both straight and curved datasets, comparing our approach with K-means and the DBSCAN pipeline proposed by \cite{mazzia2021deepway}. We first test models on our synthetic datasets (Straight Synth, Curved Synth) and then validate the results on real occupancy grids obtained from satellite images (Straight Real, Curved Real). For each test, we compare the results of models trained on straight rows with those obtained training on curved rows. We report the mean adjusted accuracy and clustering error with their standard deviations.}
\label{tab:clustering}
\vspace{5pt}
\resizebox{0.9\textwidth}{!}{%
\begin{tabular}{clclclclc}
\toprule
\textbf{Test}                &  & \textbf{Method}           &  & \textbf{Train} &  & \textbf{Adjusted Accuracy} &  & \textbf{Clustering Error} \\ \midrule \midrule
\multirow{6}{*}{Straight Synth}    &  & \multirow{2}{*}{K-means} &  & Straight       &  & \textbf{1.0000 ± 0}                 &  & \textbf{0 ± 0}                     \\
                             &  &                          &  & Curved         &  & 0.9913 ± 0.0076            &  & 0.6667 ± 0.5774           \\ \cmidrule{3-9} 
                             &  & \multirow{2}{*}{DBSCAN}  &  & Straight       &  & \textbf{1.0000 ± 0}                 &  & \textbf{0 ± 0}                     \\
                             &  &                          &  & Curved         &  & 0.9724 ± 0.0240            &  & 2.0000 ± 1.7321           \\ \cmidrule{3-9} 
                             &  & \multirow{2}{*}{Ours}     &  & Straight       &  & 0.9994 ± 0.0003            &  & 0.0187 ± 0.0114           \\
                             &  &                          &  & Curved         &  & 0.9985 ± 0.0006            &  & 0.0527 ± 0.0219           \\ \midrule
\multirow{6}{*}{Straight Real} &  & \multirow{2}{*}{K-means} &  & Straight &  & 0.4243 ± 0.1037 &  & 26.3333 ± 7.0238 \\
                             &  &                          &  & Curved         &  & 0.4635 ± 0.0873            &  & 26.0000 ± 5.1962          \\ \cmidrule{3-9} 
                             &  & \multirow{2}{*}{DBSCAN}  &  & Straight       &  & 0.9532 ± 0.0429            &  & 2.3333 ± 2.0817           \\
                             &  &                          &  & Curved         &  & 0.9585 ± 0.0026            &  & 2.0000 ± 0                \\ \cmidrule{3-9} 
                             &  & \multirow{2}{*}{Ours}     &  & Straight       &  & 0.9707 ± 0.0135            &  & 1.0400 ± 0.5197           \\
                             &  &                          &  & Curved         &  & \textbf{0.9716 ± 0.0123}            &  & \textbf{0.7700 ± 0.3012}           \\ \midrule \midrule
\multirow{6}{*}{Curved Synth}      &  & \multirow{2}{*}{K-means} &  & Straight       &  & 0.9714 ± 0.0336            &  & 1.0000 ± 1.0000           \\
                             &  &                          &  & Curved         &  & 0.9885 ± 0.0199            &  & 0.3333 ± 0.5774           \\ \cmidrule{3-9} 
                             &  & \multirow{2}{*}{DBSCAN}  &  & Straight       &  & 0.9563 ± 0.0757            &  & 1.3333 ± 2.3094           \\
                             &  &                          &  & Curved         &  & 0.8898 ± 0.0337            &  & 3.0000 ± 1.0000           \\ \cmidrule{3-9} 
                             &  & \multirow{2}{*}{Ours}     &  & Straight       &  & 0.9823 ± 0.0138            &  & 0.3414 ± 0.3278           \\
                             &  &                          &  & Curved         &  & \textbf{0.9992 ± 0.0006}            &  & \textbf{0.0127 ± 0.0038}           \\ \midrule
\multirow{6}{*}{Curved Real} &  & \multirow{2}{*}{K-means} &  & Straight       &  & 0.2443 ± 0.0984            &  & 73.3333 ± 29.2632         \\
                             &  &                          &  & Curved         &  & 0.2721 ± 0.1493            &  & 70.0000 ± 19.5192          \\ \cmidrule{3-9} 
                             &  & \multirow{2}{*}{DBSCAN}  &  & Straight       &  & 0.7247 ± 0.2734            &  & 27.0000 ± 25.5343         \\
                             &  &                          &  & Curved         &  & 0.5181 ± 0.1061            &  & 45.3333 ± 6.6583          \\ \cmidrule{3-9} 
                             &  & \multirow{2}{*}{Ours}     &  & Straight       &  & 0.8571 ± 0.0924            &  & 3.4667 ± 2.4437           \\
                             &  &                          &  & Curved         &  & \textbf{0.9344 ± 0.0116}            &  & \textbf{1.1933 ± 0.1858}           \\ \bottomrule
\end{tabular}%
}
\end{table}

The first important result is the model trained on curved crops being able to reach an $AP_8$ of about 94\% on all four test scenarios. This achievement confirms the effectiveness of our model far beyond the synthetic training scenario, as real satellite data does not seem to create substantial performance drops (5.7\% at worst). Looking at lower values of $r$, the synthetic-to-real gap rises to 11.5\%, showing how the model is able to estimate synthetic waypoints with higher precision. The model trained on straight crops achieves excellent performance on its corresponding test set and even on real satellite data, but generalizes poorly on curved rows: the precision drop reaches 11\% on $AP_8$ and even 22\% considering $AP_3$. On the contrary, the model trained on curved crops scales very well on straight scenarios. This outcome confirms the importance of training on curved crops to obtain robust models able to cope with challenging situations.

\subsection{Waypoint Clustering}

\begin{figure}[t]
\centering
\begin{subfigure}{0.32\textwidth}
  \centering
  \frame{\includegraphics[trim={50pt 25pt 20pt 20pt},clip,width=\linewidth]{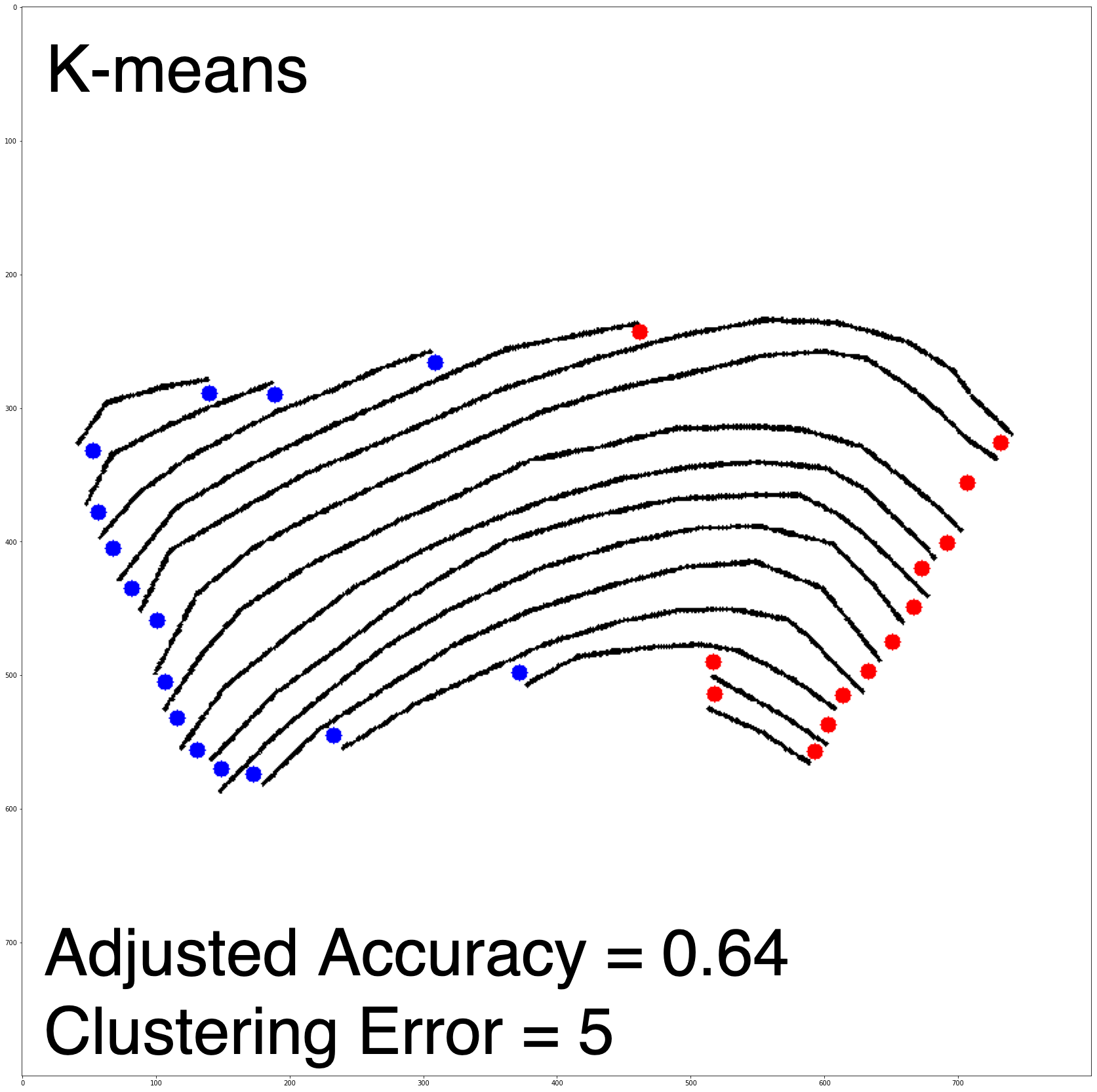}}
\end{subfigure}
\begin{subfigure}{0.32\textwidth}
  \centering
  \frame{\includegraphics[trim={50pt 25pt 20pt 20pt},clip,width=\linewidth]{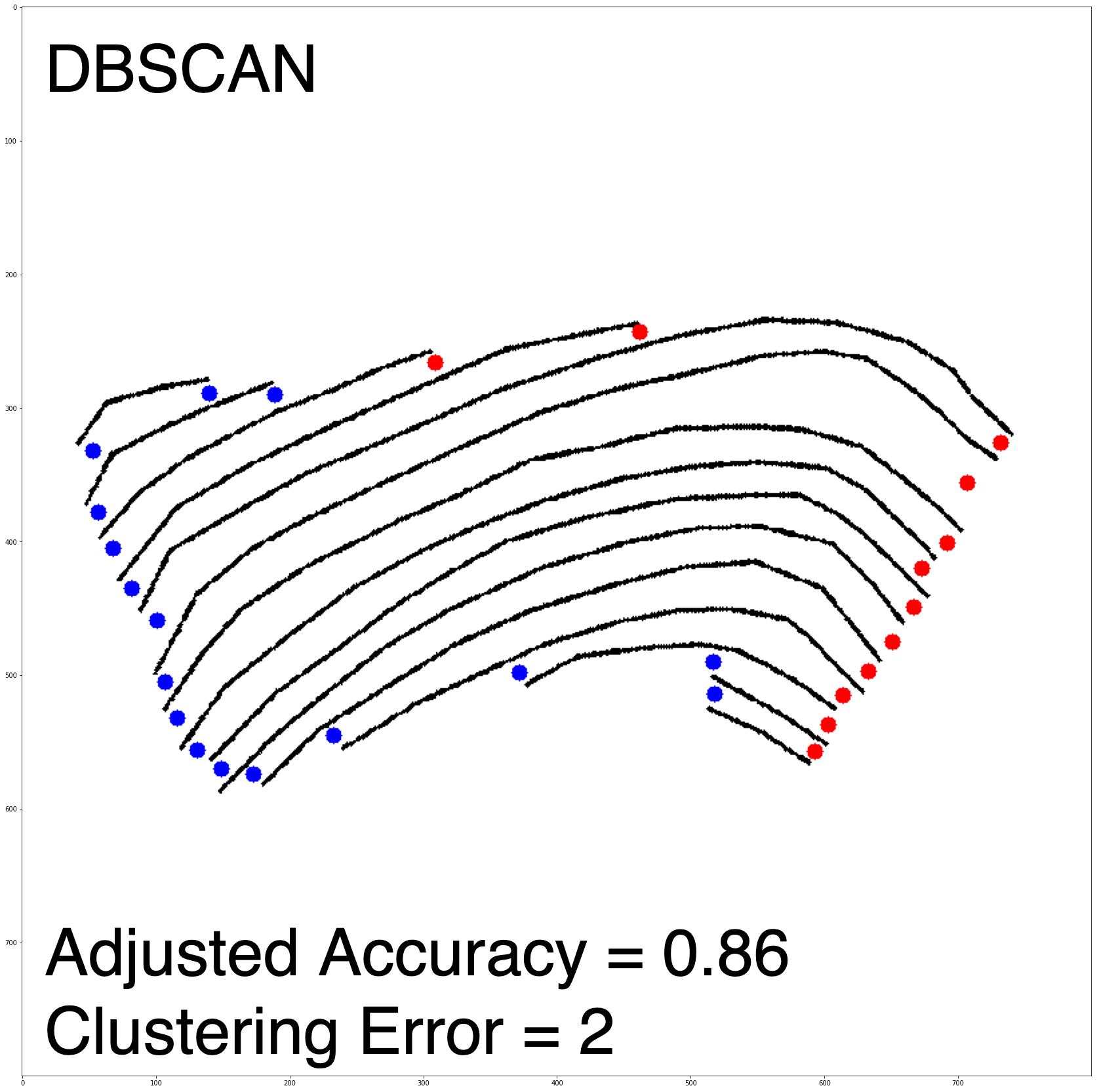}}
 \end{subfigure}
\begin{subfigure}{0.32\textwidth}
  \centering
  \frame{\includegraphics[trim={50pt 25pt 20pt 20pt},clip,width=\linewidth]{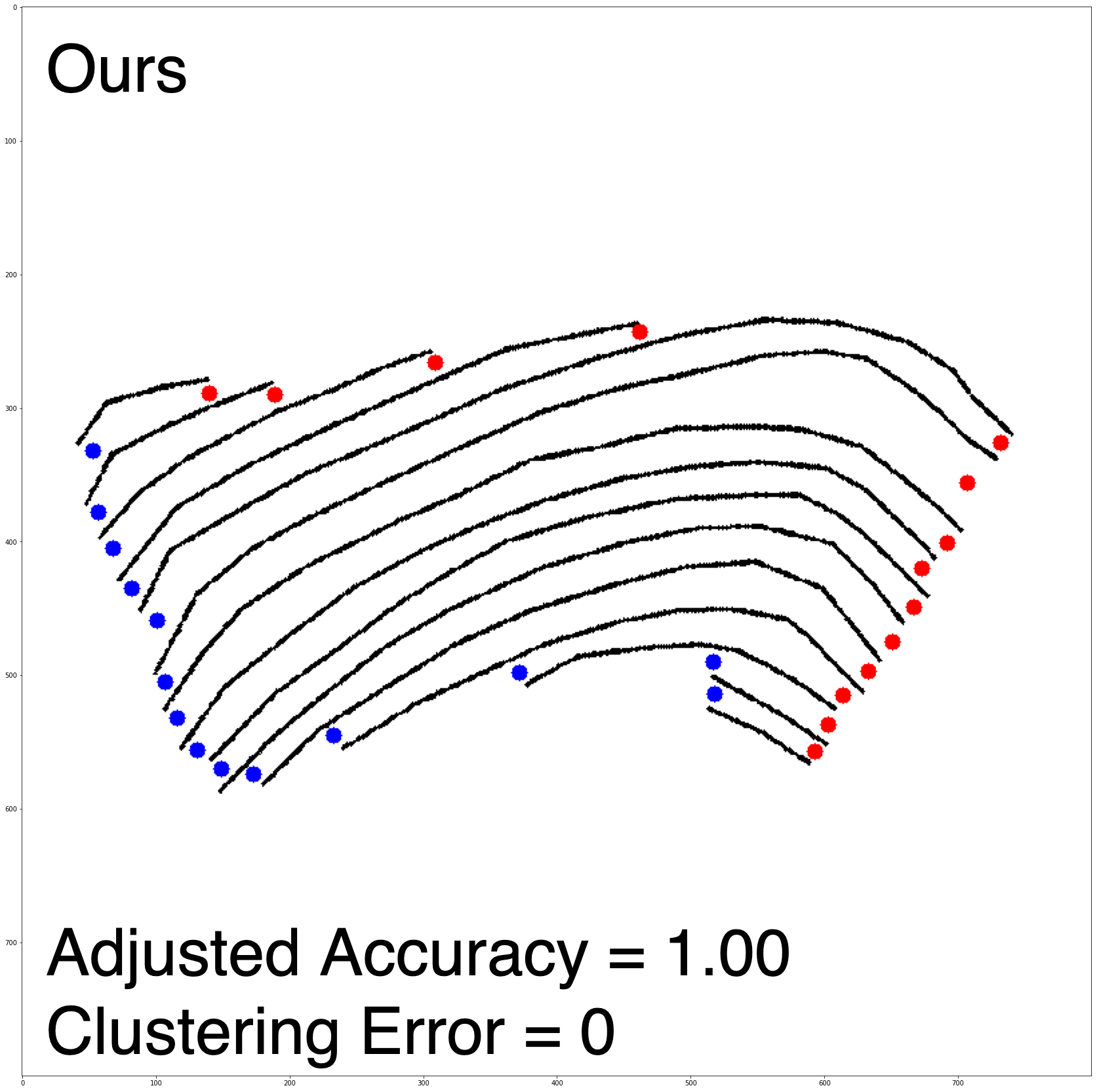}}
 \end{subfigure}
\caption{Examples of clustering on a real-world curved sample: K-means and DBSCAN pipeline \cite{mazzia2021deepway} are not able to correctly cluster the predicted waypoints; on the other hand, the proposed method correctly assigns the points.}
\label{fig:comparison}
\end{figure}

As regards waypoint clustering, we adopt two separate metrics. The first is an adjusted binary accuracy, assigning a score of 0 to the worst outcome (all the points in the same cluster, meaning 50\% of the points correctly clustered) and 1 to perfect clustering. However, the number of waypoints in a crop is variable and accuracy alone does not give an insight of the distribution of errors among different samples. For example, crops with a small number of waypoints tend to be easier to cluster than dense ones. Considering the fact that full-coverage path planning is possible only if every waypoint is correctly clustered, we add a clustering error metric computing the average number of wrongly labeled points per image. The results are detailed in Table \ref{tab:clustering}. To have a baseline, we compare our approach with the K-means algorithm directly applied in the image reference system and the DBSCAN clustering with geometrical assignment approach proposed by \cite{mazzia2021deepway}. All the clustering tests are performed setting the confidence threshold to $t_p=0.4$ and the waypoint suppression threshold to $t_{sup}=8$ pixels. As for the previous results, each value is reported with its mean and standard deviation.

Our methodology achieves remarkable results, outperforming or at least matching existing solutions in all the testing scenarios. In particular, both the training strategies (based on straight and curved crops) approach perfect clustering on the synthetic straight dataset and generalize well to real crops. On the contrary, K-means, which perfectly works for the well-separated synthetic samples, loses more than half of its adjusted accuracy and presents a very high clustering error when switching to real test rows, mainly due to the irregular shapes typical of real-world vineyards. The DBSCAN pipeline, instead, is able to generalize to straight satellite crops, since the methodology was specifically designed to cope with real-world straight rows.

As regards curved test sets, K-means clustering is totally unable to generalize to the real dataset. At the same time, also the DBSCAN pipeline results drop significantly when switching to real samples, due to its heavy dependence on angle estimation. Our model, trained on straight rows, obtains 0.98 adjusted accuracy and 0.34 clustering error on synthetic data, outperforming both the baselines. However, it struggles to generalize to real crops, reaching an adjusted accuracy of 0.86. On the other hand, the model trained on curved data outperforms the baselines in synthetic and real data, where it achieves an adjusted accuracy of 0.93. This result can be considered extremely positive, taking into account the strong challenges present in satellite data. In particular, a clustering error of 1.19 is remarkably smaller than those obtained by K-means and DBSCAN. In conclusion, these results confirm how the proposed methodology, combined with a well-devised generation process of curved synthetic samples, allows path planning even in challenging scenarios.

\begin{figure}[t]
\centering
\begin{subfigure}{0.48\textwidth}
  \centering
  \frame{\includegraphics[trim={75pt 160pt 50pt 210pt},clip,width=\linewidth]{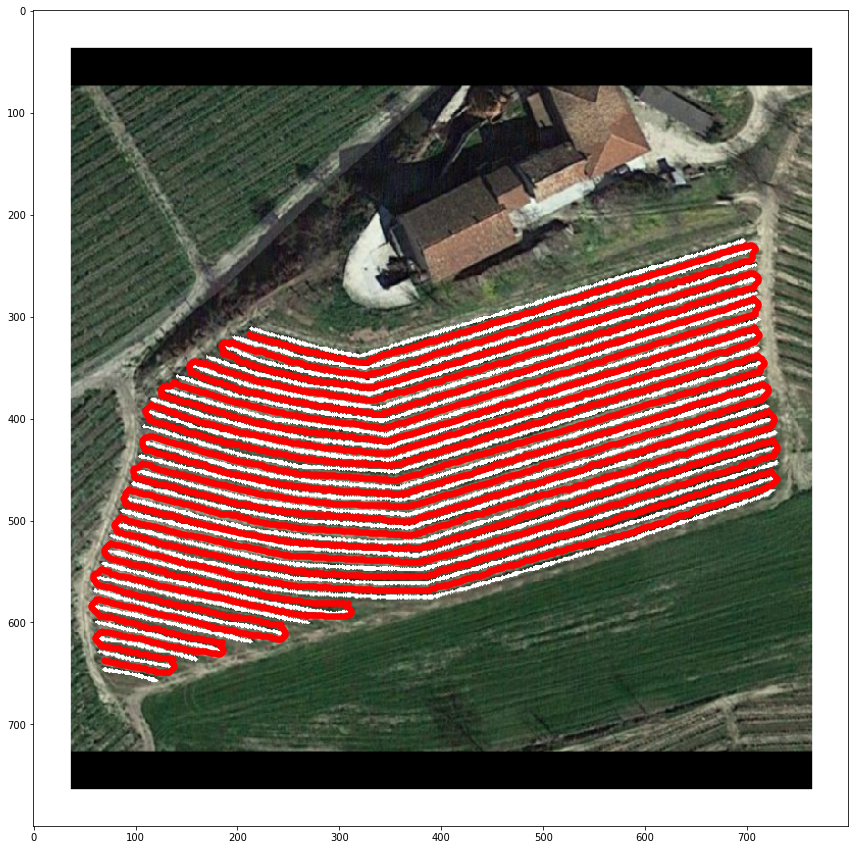}}
\end{subfigure}
\begin{subfigure}{0.48\textwidth}
  \centering
  \frame{\includegraphics[trim={75pt 185pt 50pt 185pt},clip,width=\linewidth]{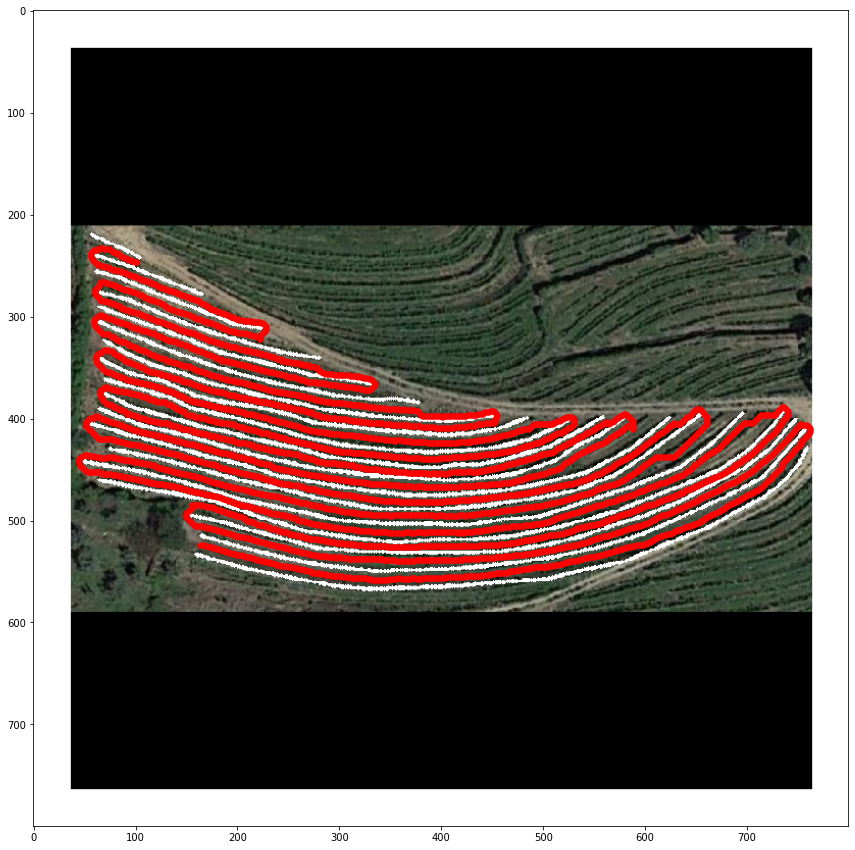}}
 \end{subfigure}
\begin{subfigure}{0.48\textwidth}
  \vspace{6pt}
  \centering
  \frame{\includegraphics[trim={75pt 200pt 50pt 170pt},clip,width=\linewidth]{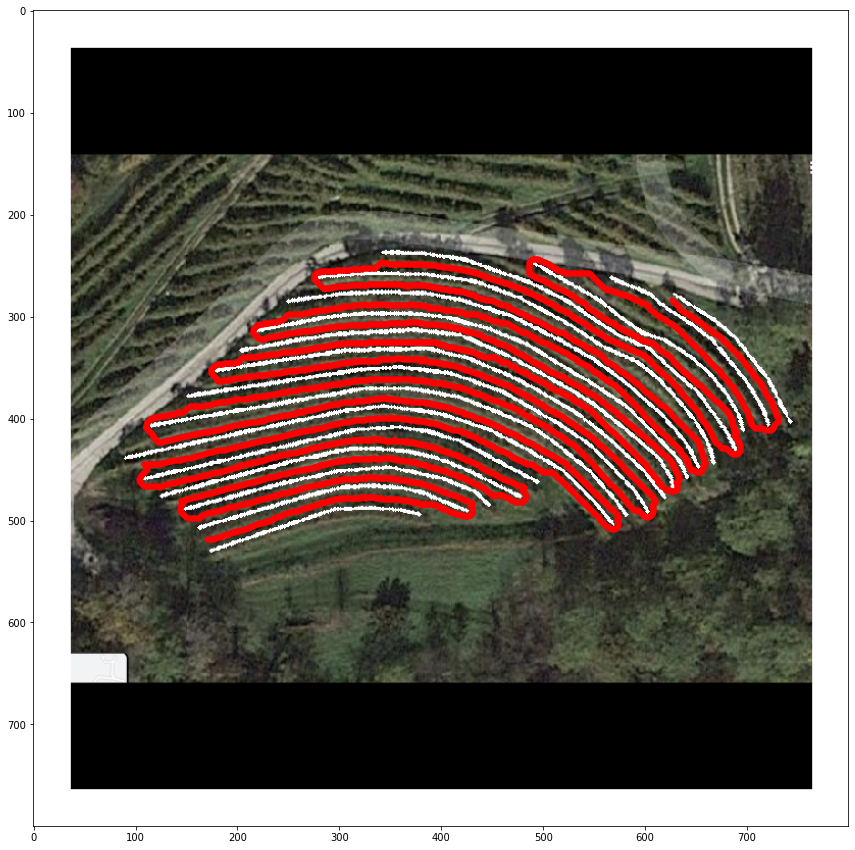}}
 \end{subfigure}
\begin{subfigure}{0.48\textwidth}
  \vspace{6pt}
  \centering
  \frame{\includegraphics[trim={75pt 185pt 50pt 185pt},clip,width=\linewidth]{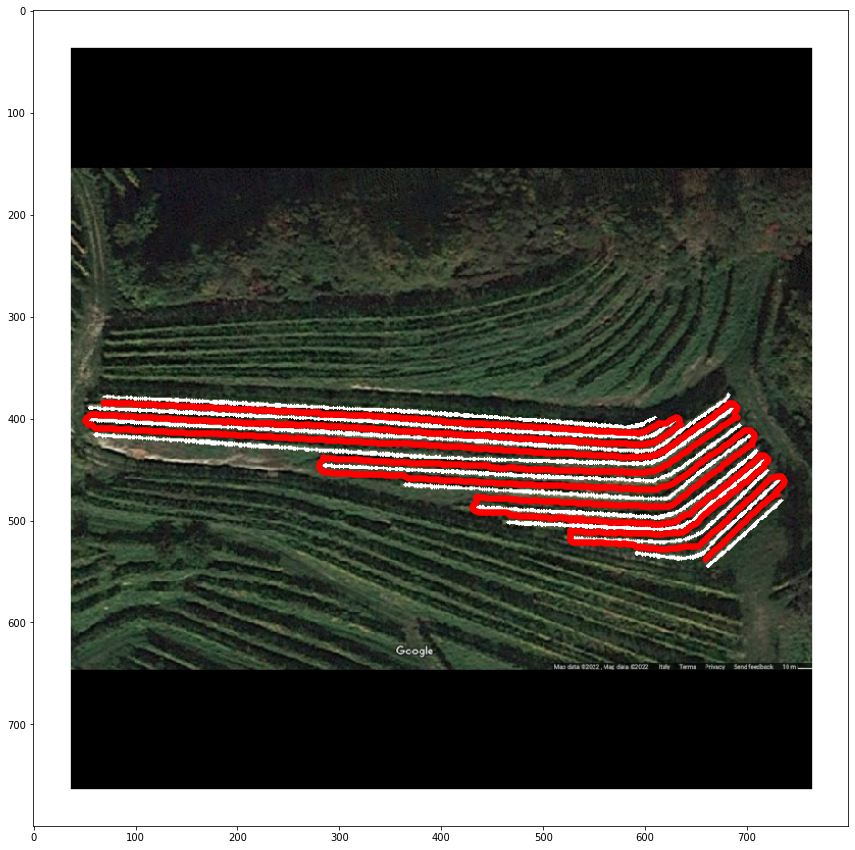}}
 \end{subfigure}
\caption{Examples of full-coverage path planning in real-world curved vineyards taken from Google Maps satellite database.}
\label{fig:path}
\end{figure}

\subsection{Qualitative Results}
To give further insight into the performance of the proposed methodology, we present some qualitative examples on real-world curved samples. Fig. \ref{fig:comparison} shows a comparison between the three clustering methodologies. K-means and the DBSCAN pipeline are clearly unable to correctly assign points in challenging scenarios. Finally, Fig. \ref{fig:path} shows some examples of full-coverage path planning. The planning is performed by selecting the points in an A-B-B-A fashion and using the planner proposed by \cite{cerrato2021adaptive}. With geo-referenced maps, the planned path can be converted from the image reference system to a Global Navigation Satellite System (GNSS) reference frame to be used in real-world navigation. All the tests are performed with the model trained on the curved dataset and setting the confidence threshold to $t_p=0.4$ and the waypoint suppression threshold to $t_{sup}=8$ pixels. 

\section{Conclusions}
\label{sec:conclusions}
In this work, we propose a novel solution for global path generation in row-based crops using deep learning and contrastive clustering. The problem of path planning in geometrically constrained environments such as vineyards and orchards has been solved through the identification of waypoints at the end of each row. Our deep learning model can simultaneously predict the position of navigation waypoints and cluster them in a unique feed-forward step. To this aim, we train the network on a synthetic dataset of top-view occupancy grids and test it on real-world satellite images, outperforming previous methodologies based on classical clustering by adopting a contrastive loss function. Our extensive experimentation demonstrates that this model successfully generalizes to challenging realistic conditions, including curved and incomplete rows.

Future works may seek the integration of the proposed waypoint generator in a complete pipeline for row-based vineyard and orchard navigation, composed of a first segmentation step to obtain the occupancy grid of the parcel from satellite or UAV imagery and a local planner for intra-row navigation.

\subsubsection{Acknowledgements} This work has been developed with the contribution of the Politecnico di Torino Interdepartmental Centre for Service Robotics (PIC4SeR) and SmartData@Polito. 

%
%
%
\bibliographystyle{splncs04}
\bibliography{bibliography}
\end{document}